\documentclass[10pt,twocolumn,letterpaper]{article}

\usepackage[pagenumbers]{iccv} 

\usepackage{algorithm}
\usepackage{algpseudocode}

\usepackage[accsupp]{axessibility} 
\usepackage{times}
\usepackage{epsfig}
\usepackage{graphicx}
\usepackage{amsmath}
\usepackage{amssymb}
\usepackage{etoc}
\usepackage{multirow}
\usepackage{color, colortbl}

\definecolor{iccvblue}{rgb}{0.21,0.49,0.74}
\usepackage[pagebackref,breaklinks,colorlinks,allcolors=iccvblue]{hyperref}

\def\paperID{3389} 
\def\confName{ICCV}
\def\confYear{2025}

\title{Towards Immersive Human-X Interaction: A Real-Time Framework for Physically Plausible Motion Synthesis}

\author{
    Kaiyang Ji$^{1}$ \qquad
    Ye Shi$^{1,2}$\qquad
    Zichen Jin$^{1}$ \qquad
    Kangyi Chen$^{1}$ \qquad
    Lan Xu$^{1,2}$ \qquad \\
    Yuexin Ma$^{1,2}$ \qquad
    Jingyi Yu$^{1,2}$ \qquad
    Jingya Wang$^{1,2,}$\thanks{Corresponding author.} \\
    $^1$ShanghaiTech University \\
    $^2$Shanghai Engineering Research Center of Intelligent Vision and Imaging\\
    {\tt\footnotesize \{jiky2024,shiye,jinzch12023,v-chenky,xulan1,mayuexin,yujingyi,wangjingya\}@shanghaitech.edu.cn} 
}

\begin{document}

\twocolumn[{
    \renewcommand\twocolumn[1][]{#1}
    \maketitle
    \vspace{-1cm}
    \begin{center}
        \includegraphics[width=1.0\linewidth]{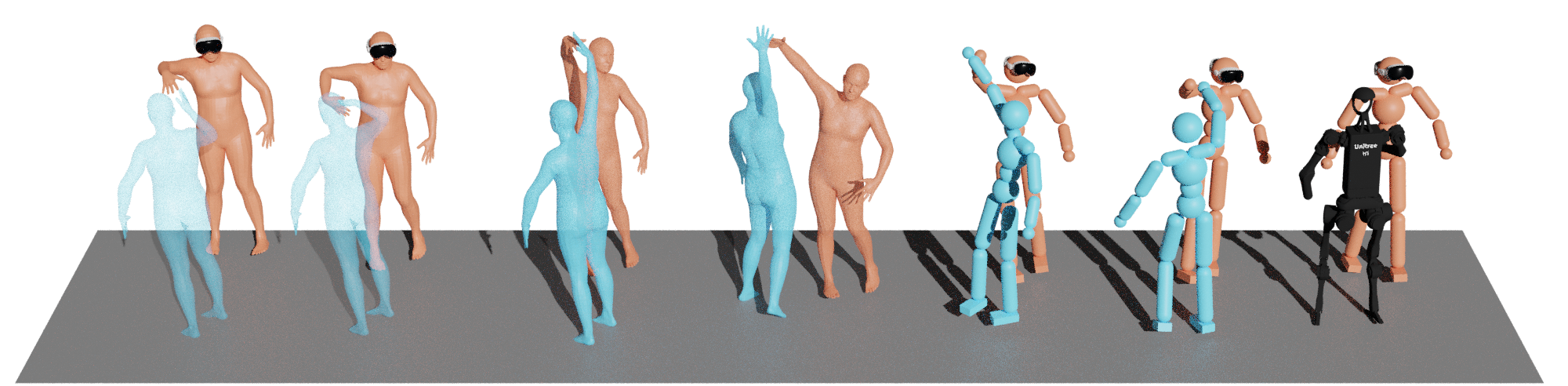}
        \captionof{figure}{We propose Human-X, the first framework designed to enable latency-free interaction between humans and diverse entities, including human-avatar, human-humanoid, and human-robot interaction. }

        \label{fig:main}
    \end{center}
}]

\def\thefootnote{*}\footnotetext{Corresponding author.}

\begin{abstract}
Real-time synthesis of physically plausible human interactions remains a critical challenge for immersive VR/AR systems and humanoid robotics. While existing methods demonstrate progress in kinematic motion generation, they often fail to address the fundamental tension between real-time responsiveness, physical feasibility, and safety requirements in dynamic human-machine interactions. We introduce Human-X, a novel framework designed to enable immersive and physically plausible human interactions across diverse entities, including human-avatar, human-humanoid, and human-robot systems. Unlike existing approaches that focus on post-hoc alignment or simplified physics, our method jointly predicts actions and reactions in real-time using an auto-regressive reaction diffusion planner, ensuring seamless synchronization and context-aware responses. To enhance physical realism and safety, we integrate an actor-aware motion tracking policy trained with reinforcement learning, which dynamically adapts to interaction partners’ movements while avoiding artifacts like foot sliding and penetration. Extensive experiments on the Inter-X and InterHuman datasets demonstrate significant improvements in motion quality, interaction continuity, and physical plausibility over state-of-the-art methods. Our framework is validated in real-world applications, including virtual reality interface for human-robot interaction, showcasing its potential for advancing human-robot collaboration. Project page: {\url{https://humanx-interaction.github.io/}}
\end{abstract} 

\etocdepthtag.toc{mtchapter}
\etocsettagdepth{mtchapter}{subsection}
\etocsettagdepth{mtappendix}{none}

\section{Introduction}
\label{sec:intro}

\begin{table*}[htbp]
    \resizebox{1.0 \textwidth}{!}{
    \begin{tabular}{c|ccccccc}
        \toprule
        Methods & physics-based & immersive inference & latency-free & real-time model & generation ability & strong contact & text-control \\
        \midrule
        InterGen \cite{liang2024intergen} &  &  &  &  & \checkmark & \checkmark & \checkmark  \\
        ReGenNet \cite{xu2024regennet} &  &  &  & \checkmark & \checkmark & \checkmark & \checkmark \\
        SoLaMi \cite{jiang2024solami} &  & \checkmark &  &  & \checkmark &  & \checkmark \\
        PhysReaction \cite{liu2024physreaction} & \checkmark &  & \checkmark & \checkmark &  & \checkmark &  \\
        Think-Then-React \cite{tanthink} &  &  &  & \checkmark & \checkmark & \checkmark & \checkmark \\
        Human-X & \checkmark & \checkmark & \checkmark & \checkmark & \checkmark & \checkmark & \checkmark \\
        \bottomrule
    \end{tabular}}
    \caption{Comparative analysis of Human-X versus existing human reaction synthesis approaches.}
    \label{tab:task-com}
\end{table*}

Human-human interaction synthesis has long been a focal point in the field of motion generation, with widespread applications in social dynamics, gaming, and animation \cite{jiang2024solami, starke2020local}.  With the rapid development of virtual reality (VR) and augmented reality (AR), these technologies now empower the creation of deeply immersive and contextually rich interactions between humans and digital avatars. Particularly, the emergence of humanoid robots further bridges the gap between virtual and physical realms, enabling robots to engage with humans in daily life through adaptive and socially aware behaviors. However, achieving real-time prediction of diverse and safety-guaranteed humanoid reactions, an essential capability for applications ranging from daily companionship to assistive care, remains a critical, unsolved challenge.

Current approaches to human interaction modeling rely heavily on offline training data or simplified physics assumptions, leading to limited adaptability in dynamic scenarios. While data-driven methods such as \cite{liang2024intergen, xu2024inter, javed2024intermask} have shown promise in synthesizing kinematic motions interaction, they often fail to account for the intricate physical constraints and safety requirements inherent in human-humanoid cohabitation. 
For instance, abrupt or unstable motions generated through these techniques pose risks in shared environments. This gap underscores the need for a unified interaction framework that integrates real-time reactivity, physical realism, and safety assurance.


In this work, we explore Human-X Interaction (Fig.\ref{fig:main}), a systematic study of immersive interactions across three domains: 1) Human-Avatar Interaction using kinematic-driven motion in VR/AR; 2) Human-Humanoid Interaction with physics-based motion synthesis; 3) Human-Robot Interaction ensuring safe physical contact. Unlike existing physics-aware methods limited to replicating training data, our framework balances natural movement with physical feasibility while maintaining real-time performance. This integration enables safer and more natural human-machine collaboration across applications from entertainment to assistive robotics. Achieving real-time synchronization between interacting partners faces major technical challenges. Conventional methods rely exclusively on past actor motions to generate reactions, leading to temporal mismatches. Furthermore, ensuring physical realism is more challenging, as existing solutions focus mainly on basic collision avoidance and neglect dynamic interaction forces.

To overcome these limitations, we present Human-X, an Auto-Regressive Action-Reaction Diffusion Planner for real-time interaction. Our framework jointly models both partners' motion histories to predict subsequent actions and reactions. We introduce multiple contact-aware losses to train realistic interactions, and combine the diffusion model with an actor-aware tracking policy to ensure physical feasibility in humanoid/robot scenarios. In summary, our contributions can be summarized as follows: 
\begin{itemize}
    \item The first auto-regressive diffusion model that jointly predicts actions and reactions using both partners' motion history with comprehensive contract losses to ensure real-time, immersive and realistic motion synthesis. 
    \item An actor-aware reaction policy trained with reinforcement learning that dynamically adapts to interaction partners' motions, preventing common artifacts like foot sliding while maintaining physical feasibility.
    \item A complete real-time VR system demonstrating practical effectiveness. Experiments on Inter-X and InterHuman datasets show significant improvements in motion quality, interaction continuity, and physical accuracy compared to existing methods. 
\end{itemize}

\section{Related Works}
\label{sec:rel_works}

\begin{figure*}[t]
  \centering
    \includegraphics[width=1.0\linewidth]{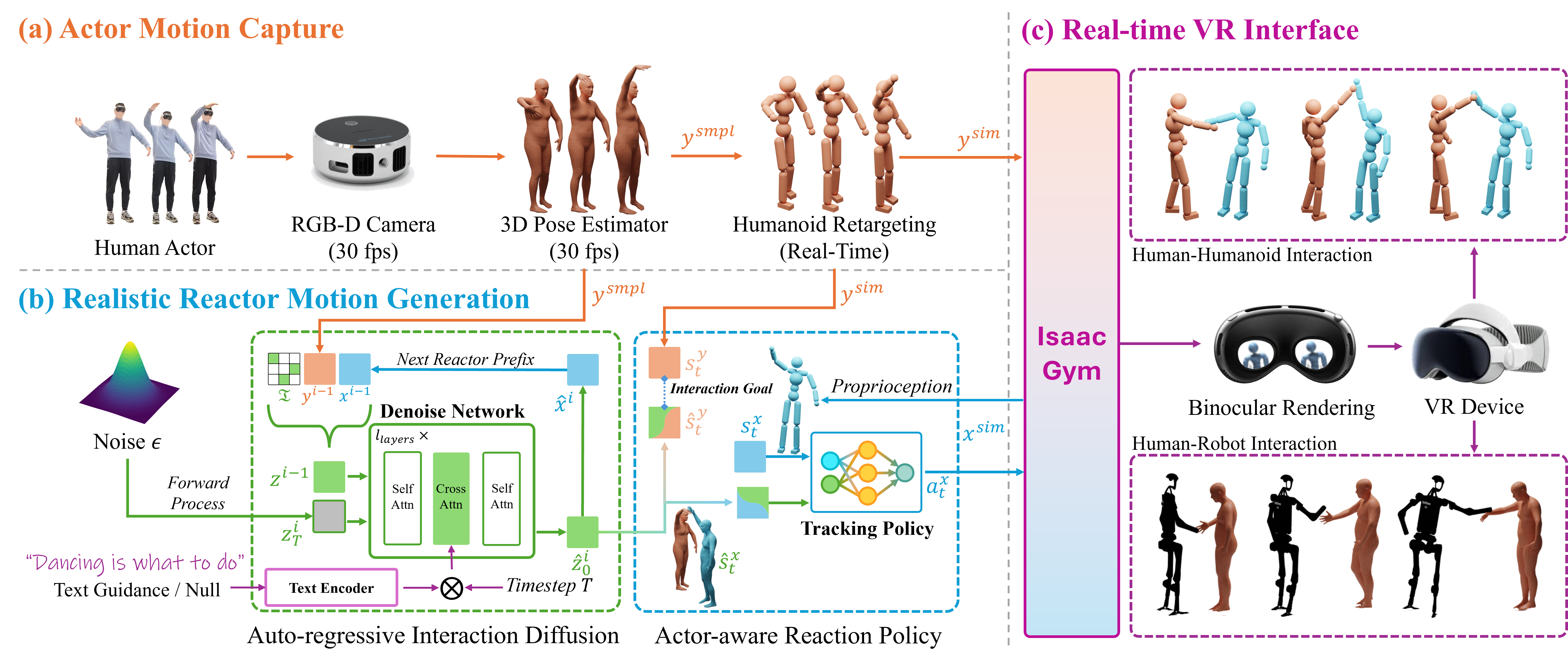}

   \caption{ Overview of our immersive real-time interaction synthesis pipeline:
\textbf{(a) Actor Motion Capture}: A human actor’s movements are recorded at 30 fps by an RGB-D camera and translated into 3D poses, which are then retargeted to a humanoid character.
\textbf{(b) Realistic Reactor Motion Generation}: An auto-regressive diffusion model, guided by optional text prompts (e.g., “Dancing is what to do”), generates plausible reaction motions. These motions are tracked by an actor-aware controller, which uses proprioception signals to ensure realistic, synchronized interactions. \textbf{(c) Real-time VR Interface}: The generated and tracked motions are rendered in simulator, providing both a third-person view and a binocular VR view.}
   \label{fig:pipeline}
\end{figure*}

\subsection{Human Motion Generation}
Human-centric motion generation has seen remarkable advancements in recent years, driven by breakthroughs in large language models \cite{achiam2023gpt, jiang2023motiongpt} and generative frameworks \cite{goodfellow2014generative, ho2020denoising, kingma2013auto, yang2024guidance, zhu2025unidb}. Notably, the introduction of motion diffusion models \cite{tevet2023human, zhang2024motiondiffuse} has significantly elevated the performance of single-person motion generation. Building on these developments, researchers have increasingly focused on enriching interactions with the environment by incorporating diverse modalities, such as text \cite{wu2024human, wu2024thor, jiang2023motiongpt,zhang2023remodiffuse}, audio \cite{alexanderson2023listen, tseng2023edge}, and scene information \cite{du2023avatars, huang2023diffusion, li2023ego, tang2024unified}. For instance, recent works \cite{wu2024human, wu2024thor, deng2025human, tian2024gaze, zhang2025openhoi} have leveraged text-based guidance to facilitate Human-Object Interactions (HOI), while MotionGPT \cite{jiang2023motiongpt} treats motion as a form of language, enabling seamless bidirectional translation between text and motion. Despite these advancements, existing approaches face a critical limitation: their model architectures are inherently ill-suited for real-time reaction synthesis.

\subsection{Human-Human Interaction}
Research on human-human interaction synthesis has evolved around three primary directions: (1) Interactive Behavior Modeling leverages actor-reactor role decomposition to capture interaction dynamics. For example, \cite{tanaka2023role} identifies role-specific patterns, while \cite{xu2023actformer, liang2024intergen} use transformer architectures to model multi-agent dependencies. (2) Multimodal Interaction integrates text, audio, or scene context to enhance realism \cite{li2024interdance, siyao2024duolando}, but often overlook dynamic physical constraints critical for real-world deployment. (3) Reactive Motion Generation employs sequence models \cite{liu2023interactive} and diffusion models \cite{xu2024regennet, jiang2025arflow} to synthesize temporally coherent reactions. These methods, however, struggle with physical plausibility and temporal consistency in long-horizon interactions. Existing Reactive Generation methods face a fundamental tension: Human interaction models achieve reactivity but lack physical grounding, while physics-based animations enforce plausibility but are agnostic to interaction dynamics. While PhysReaction \cite{liu2024physreaction} introduced the first physics-based approach for reaction synthesis using imitation learning, it was limited to mimicking known distributions, lacking generalization ability, diversity, controllability, and the capacity to handle unseen tasks.
Our work bridges this gap by unifying interaction-aware reactivity with physically constrained control.

\subsection{Physics-based Motion Synthesis}
In contrast to kinematics-based methods, physics-based motion synthesis has emerged as a widely adopted approach for generating realistic and physically plausible character movements by integrating physical principles into the animation process. Early work, such as DeepMimic \cite{peng2018deepmimic}, leveraged deep reinforcement learning to enable characters to imitate diverse motion clips within a physics-based framework. Subsequent advancements \cite{ho2016generative, peng2021amp} utilized generative adversarial networks to enhance stylized motion realism. Progress continued with PHC \cite{luo2023perpetual} and PULSE \cite{luo2023universal}, focusing on improving the diversity of generated motions. A significant milestone was achieved with PhysDiff \cite{yuan2023physdiff}, which introduced physics-guided diffusion models to mitigate artifacts such as floating and foot sliding.
Despite these advancements, the physical plausibility of human-human interactions remains largely unexplored. 
\section{Method}
\label{sec:method}

\subsection{System Overview}
\label{subsec:overview}
In this section, we present a detailed overview of our immersive interaction synthesis system. To facilitate realistic and responsive interactions between two agents in immersive environments, our system takes online actor motion capture data as input, processes it through an advanced motion generation pipeline, and outputs continuous, reactive motion in real-time, as shown in Fig.\ref{fig:pipeline}. The first key component of our system, detailed in Sec.\ref{subsec:pose}, is the real-time capture of the actor's pose and the reconstruction of its state in physics simulation. Subsequently, the Auto-regressive Interaction Diffusion Model, as described in Sec.\ref{subsec:diffusion}, is employed to generate the reaction of one agent in response to the actor context. Next, we introduce the Reaction Motion Tracker in Sec.\ref{subsec:phc}, which ensures that the motion of the reactor adheres to physical constraints for realism and natural behavior. Finally, we explain how VR Interface works in Sec.\ref{subsec:vr}, which enables immersive, real-time interaction with the generated motions. 

\subsection{Human Pose Estimation and Reconstruction}
\label{subsec:pose}
We firstly utilize a real-time RGB-D camera to capture images. Then, HybrIK \cite{li2025hybrik} is employed for latency-free reconstruction of the human pose. However, due to the inaccuracies in the depth information provided by HybrIK, the depth map is directly obtained from the RGB-D camera instead. Following \cite{luo2023perpetual,luo2024real}, we bridge SMPL's kinematic skeleton to physics simulation via applying a simple retargeting method. We utilize Isaac Gym \cite{makoviychuk2021isaac} as our physics simulation platform to create a third-person perspective for the interaction between the actor and reactor. Besides, the image reading process, motion generation of the reactor, and the physical simulation within Isaac Gym are handled by separate processes. ROS2 is used as the communication framework to facilitate real-time data transmission.

\subsection{Online Human Action-Reaction Synthesis}
\label{subsec:diffusion}

The goal of online action-reaction synthesis is to generate realistic and continuous future $k$ frames of reactor motions $\mathbf{x}^i$ := $\{x^{n:n+k-1}\}$ given $h$ history frames of interaction contexts $\{\mathbf{x}^{i-1}, \mathbf{y}^{i-1}\}$ := $\{x^{n-h: n-1}, y^{n-h: n-1}\}$ at current frame number $n$ and current window index $i$. During each window prediction, an optional text prompt $c$ can be added to guide the reaction generation process. 

\subsubsection{Reactor-centric Interaction Modeling}
\label{subsubsec:rep}

\paragraph{Reactor-centric Canonicalization.}
For individual motion synthesis, the initial frame of each sequence is typically aligned at the root to improve the generalizability of the model \cite{guo2022generating, tevet2024closd}. However, using the same normalization for human-human interaction data can eliminate key details about the mutual spatial relationship. To retain relative positioning without losing generalization ability, we leverage a reactor-centric canonicalization that the pose of reactor is realigned at the root and oriented toward the forward direction, while the pose of actor is adjusted by the same transformation. This method maintains the essential spatial context between the two, enabling the model to accurately learn their interdependent movement dynamics.

\paragraph{Reactor-centric Interaction Representation.} We model each frame of interaction as a hybrid representation of both actor and reactor motion. For reaction representation, we follow previous works \cite{tevet2023human, guo2022generating} and use the HumanML3D representation. The reactor motion $x$ in frame $n$ is defined by $x^n = (r^x, \dot{r}^x, p^x, \dot{p}^x, \theta^x, f^x)$, where $r^x \in \mathbb{R}$ denotes the height of the roots, $\dot{r}^x \in \mathbb{R}^3$ denotes the angular velocity of the roots along the Y axis and the linear velocity of the roots in the XZ plane, $p^x \in \mathbb{R}^{(J - 1) \times 3}$ are the local joint positions, $\dot{p}^x \in \mathbb{R}^{J \times 3}$ denotes the temporal difference of the local joint positions, $\theta^x \in \mathbb{R}^{(J - 1) \times 6}$ is the 6D representation \cite{zhou2019continuity} of the joint rotations and $f^x \in \mathbb{R}^{4}$ are binary labels for the detection of contact. For actor motion captured from pose estimation, we define the actor $y$ in frame $n$ by $y^n =(r^{x\to y}, \dot{r}^{x\to y}, p^{x\to y}, \dot{p}^{x \to y}, f^y)$ where all the features are in the reactor's reference frame, making it a pure relative representation consisting of relative root transition, relative root rotation, relative root linear velocity, relative joint positions, relative joint velocity and feet contact labels. 
To emphasize actor-reactor contacts, we introduce the binary interaction field $\mathcal{I}^{n} \in \mathbb{R}^{\bar{J} \times \bar{J}}$ obtained by thresholding the relative distance between the selected $\bar{J}$ joints of each agent in a contact map. For body model selection, we follow the skeleton of SMPL \cite{loper2015smpl} with $J = 22$ joints. The total reactor-centric interaction $\mathbf{z}$ at frame $n$ representation can be formed as, 
\begin{equation}
z^n = (x^n, y^n, \mathcal{I}^n),
\end{equation}
with $x^n$ denotes the reactor motion, $y^n$ denotes the relative actor motion and $\mathcal{I}^n$ is the joint contact features. Thus, we can model the interaction sequence as $\mathbf{z}^i= \{z^{n:n+k-1}\}$ given $k$ frames and window index $i$. 

\subsubsection{Auto-regressive Interaction Diffusion Planner}

Based on the interaction motion representation provided in Sec.\ref{subsubsec:rep}, we transfer raw motion data into a canonical space suitable for generative learning with denoising networks. For probabilistic generator, we follow the DDPM \cite{ho2020denoising} framework by incrementally adding Gaussian noise to the clean data sample $z_0$ over $T$ steps: 
\begin{equation}
    q(\mathbf{z}_t|\mathbf{z}_{t-1}) = \mathcal{N}(\sqrt{\alpha_t}\mathbf{z}_{t-1}, (1-\alpha_t)\mathbf{I}),
\end{equation}
where $\alpha_t \in (0, 1)$ are constant parameters for variance scheduling. Given the noised data $\mathbf{z}_t^i$ at the $i$-th window clip, the diffusion step $t$, the prefix interaction window $\mathbf{z}^{i-1} = z^{n-h:n-1}$, and the optional text prompt $c$, we train a transformer-based denoiser $\mathcal{G}$ to predict the cleaned interaction primitive itself instead of predicting the noise \cite{tevet2023human}:
\begin{equation}
    \hat{\mathbf{z}}_0^i = \mathcal{G}(\mathbf{z}^i_t, \mathbf{z}^{i-1}, t, c),
\end{equation}
where the text prompt $c$ is randomly masked by a rate of $0.15$ during training to enable unconditioned generation and classifier-free guidance \cite{ho2022classifier}. The predicted motion $\hat{\mathbf{z}}_0^i$ is then inserted to the interaction history for next interaction window extraction after the diffusion sampling loop \cite{ho2020denoising, song2020denoising}.

\paragraph{Losses}
The denoised motion $\hat{\mathbf{z}}_0^i$ is supervised by a simple objective in DDPM \cite{ho2020denoising}: 
\begin{equation}
    \mathcal{L}_{simple} = \mathbb{E}_{\mathbf{z}^i_0 \sim q(\mathbf{z}^i_0), t \sim [1, T]} [ \| \mathbf{z}_0^i - \hat{\mathbf{z}}_0^i \|_2^2].
\end{equation}
We extend the training objective with several auxiliary losses to enhance interaction quality and realism. To optimize foot contact quality, we add a foot contact loss on both agents:
\begin{equation}
    \mathcal{L}_{foot} =  \sum_{j \in \{x^{feet}, y^{feet}\}}\|\dot{p}^j(\hat{\mathbf{z}}_0^i) \odot f^j(\hat{\mathbf{z}}_0^i)  \|_2^2,
\end{equation}
where $\dot{p}^j(\hat{\mathbf{z}}_0^i)$ is the corresponding foot joints velocity, $f^j(\hat{\mathbf{z}}_0^i)$ are foot contact label described in Sec \ref{subsubsec:rep} and $\odot$ denotes for Hadamard product. Note that $\dot{p}^j(\hat{\mathbf{z}}_0^i)$  is defined by the relative local velocity in the reactor frame, and we transform it to the global velocity in loss calculation. For interaction refinement, we design an interaction loss on contact joints distance:
\begin{equation}
    \mathcal{L}_{inter} = \sum_{x \in \bar{J}^x, y \in \bar{J}^y}\| (p^x(\hat{\mathbf{z}}_0^i)  - p^y(\hat{\mathbf{z}}_0^i)) \odot \mathcal{I}^{x,y}(\hat{\mathbf{z}}_0^i)\|_2^2,
\end{equation}
where $\mathcal{I}^{x,y}(\hat{\mathbf{z}}_0^i)$ are binary body interaction fields and $p^x(\hat{\mathbf{z}}_0^i)$, $p^y(\hat{\mathbf{z}}_0^i)$ are contact joints positions in the same root frame to enhance strong interaction. To improve the continuity of auto-regressive generation, we finally add a prefix loss $\mathcal{L}_{prefix}$ to evaluate the difference between the end of prefix motion and the beginning of the generated motion. The full loss applied to the model is then:
\begin{equation}
\begin{aligned}
    \mathcal{L} = \mathcal{L}_{simple} &+ \lambda_{foot}\mathcal{L}_{foot} 
    +\lambda_{inter}\mathcal{L}_{inter}\\
    &+ \lambda_{prefix}\mathcal{L}_{prefix},
\end{aligned}
\end{equation}
where $\lambda_{foot}$, $\lambda_{inter}$ and $\lambda_{prefix}$ are hyper-parameters to balance the loss terms from different scaling aspects.

\paragraph{Training}
During training, we utilize the scheduled training strategy \cite{bengio2015scheduled,rempe2021humor,yuan2023learning,zhao2024dart} to enhance auto-regressive generation stability and text prompt controllability. The motion is cropped to sequences of $N$ consecutive interaction windows with a random offset from the dataset. During training, $\mathbf{z}_0^i$ is noised to $\mathbf{z}_t^i$ at diffusion timestep $t$, and interaction history $\mathbf{z}^{i-1}$ is divide into 3 stage with the sample probability $p$: ground truth for supervised learning ($p=1$), random sampling for scheduled training (decaying $p$), and auto-regressive prediction ($p=0$). Details in Appendix \ref{sup:train}.

\paragraph{Inference}

\begin{algorithm}[ht]
\caption{Online reaction sampling using auto-regressive motion diffusion model}
\label{alg:sampling}
\begin{algorithmic}
\algrenewcommand\algorithmiccomment[1]{\hfill\(\triangleright\) #1}
    \State \textbf{Input:} denoiser $\mathcal{G}$, online actor loader $\mathbf{Y}$, online text loader $\mathbf{C}$, total diffusion steps $T$, classifier-free guidance scale $w$, auto-regressive reaction sampler $\mathcal{S}$.
    \State \textbf{Output:} reactor motion sequence $\mathbf{X}$
    \State $\mathbf{z}_T^0 \gets \mathcal{N}(0, 1)$
    \State $\mathbf{X} \gets \mathcal{S}(\mathcal{G}, \mathbf{z}_T^0, T,  \o, \o, w)$ \\
    {\color{gray}\Comment{init with unconditioned sampling}}
    \For{$i \gets 1$ \textbf{to} N} {\color{gray}\Comment{num of actor windows}}
        \State $\mathbf{y}^{i-1} \gets \Call{Collect}{\mathbf{Y}}, \mathbf{x}^{i-1} \gets \Call{Collect}{\mathbf{X}}$
        \State $\mathbf{z}^{i -1} \gets \Call{Canonicalize}{\mathbf{x}^{i-1}, \mathbf{y}^{i-1}}$ \\
        {\color{gray}\Comment{get interaction history}}
        \State $\mathbf{z}_T^i \gets \mathcal{N}(0, 1)$
        \State $c^i \gets \Call{Collect}{\mathbf{C}}$ 
        \State $\hat{\mathbf{z}}_0^i \gets \mathcal{S}(\mathcal{G}, \mathbf{z}_T^i, T, \mathbf{z}^{i-1}, c^i, w$) 
        {\color{gray}\Comment{CFG sampling loop}}
        \State $\hat{\mathbf{x}}^i \gets\Call{Recover}{\hat{\mathbf{z}}_0^i}$ 
        \State $\mathbf{X} \gets \mathbf{X} \cup \hat{\mathbf{x}}^i$  \\
        {\color{gray}\Comment{insert generated reaction to reactor sequences}}
    \EndFor
    \State return $\mathbf{X}$
\end{algorithmic}
\end{algorithm}

After training a denoise network $\mathcal{G}$, we can auto-regressively generate reaction sequences given interaction history $\mathbf{z}^{i-1}$, a diffusion sampler $\mathcal{S}$ like \cite{ho2020denoising, song2020denoising} and online text guidance $c$ by sampling with classifier-free guidance \cite{ho2022classifier}:
\begin{equation}
\begin{aligned}
    \mathcal{G}_w(\mathbf{z}_t^i, &\mathbf{z}^{i-1}, t, c) =  \mathcal{G}(\mathbf{z}_t^i,\mathbf{z}^{i-1}, t, \o)\\
    + \quad& w \cdot \Big(\mathcal{G}(\mathbf{z}_t^i,\mathbf{z}^{i-1}, t, c) - \mathcal{G}(\mathbf{z}_t^i,\mathbf{z}^{i-1}, t, \o)\Big),
\end{aligned}
\end{equation}
where $w$ is the guidance hyper-parameter, as shown in Alg.\ref{alg:sampling}.

\subsection{Reaction Policy}
\label{subsec:phc}
Our actor-motion data captured from virtual reality or existing datasets roughly approximates the dynamic plausibility. In contrast, the generated reactor motions is not always lying in a dynamically consistent manifold. Under the purpose of building the action-reaction controller in a physical world, we leverage the perpetual humanoid controller (PHC) \cite{luo2023perpetual}, a universal physics tracker to track the generated reaction.

PHC follows the conventional pipeline of generative adversarial and goal-conditioned reinforcement learning \cite{liu2022goalconditionedreinforcementlearningproblems, ho2016generative}, it receives target joint positions or rigid body’s keypoints as the policy goals and outputs physically retargeted motions. The policy consists of multiple actor-critic model in a multiplicative manner. Given arbitrary action-reaction generation, we formulate the auto-regressive interaction as a standard Markov Decision Process (MDP), denoted as tuple $\langle \mathcal{S}, \mathcal{A}, \mathcal{P},  r, \gamma, \rho_0 \rangle$, representing state space, action space, transition distribution, reward, discount factor and initial states' distribution. 
In our settings, the MDP is also conditioned on action-reaction tuple $\hat{z}$, the standard bellman equation of state-action value can be reformulated as: 
\begin{equation}
\begin{aligned}
Q^\pi(s,a,\hat{z}) = \mathbb{E}_{s' \sim \mathcal{P}(\cdot|s,a,\hat{z})}
\big[ r(s,a,s',\hat{z}  )+ \\
\gamma \, \mathbb{E}_{a' \sim \pi(\cdot|s',\hat{z})} [ Q^\pi(s',a',\hat{z}' ] 
\,\big],\
\end{aligned}
\end{equation}
where $\hat{z}= \langle\hat{x}, \hat{y}, y_{real} \rangle$ is the augmented interaction goal, $\hat{x}$ means a single predicted reactor frame regarded as the target goal, $\hat{y}$ and $y_{real}$ are the actor motions coming from the diffusion model generation and real world inputs. 

\paragraph{Actor-aware Reaction Policy}

To conserve computational and communication resources while prioritizing interaction safety, we integrate a low-frequency (0.5 Hz) diffusion planner with a high-frequency (30 Hz) physics tracker and real-world actor capture interface. This frequency mismatch can cause the reaction policy to pursue stale targets, elevating collision risk when the actor’s motion deviates unexpectedly. To safeguard against interpenetration, we incorporate both generated and captured actor motion histories into the policy observation and heavily penalize the trajectory inconsistency in the reward function, thereby enforcing collision-free, safety-critical responses.
More details of implementations in Appendix \ref{sup:policy-train}.

\section{Applications}
\subsection{Immersive VR Interface}
\label{subsec:vr}
Our system is built upon Unity, enabling real-time interaction and visualization in a VR environment. Leveraging VisionPro \cite{park2024using}, we develop a VR platform that allows the user to wear a headset and engage in immersive, egocentric interactions with the character. The camera in Unity is positioned on the actor's head, ensuring that it moves and rotates with the actor's motions. It continuously streams real-time simulations from Unity to the VR headset, providing an immersive experience. The user can perform any actions to guide the character's movement, such as completing a duet dance with the character.

\subsection{Human-Robot Interaction Interface}
\label{subsec:robot}
Benefiting from our robust Reaction Motion Tracker, the generated motions adhere to physical constraints and exhibit high adaptability. As a result, we can seamlessly transfer physically plausible humanoid interaction to human-robot interaction once we retarget the imitation goal from SMPL\cite{loper2015smpl} joints to specific robot joints. An optimization-based method is leveraged following \cite{ji2024exbody2, he2024learninghumantohumanoidrealtimewholebody, he2024omnih2o}. First  we fit the SMPL shape parameters $\beta$ and use $\beta$ to solve for optimal robot joint angles $\theta$. Then the corresponding joints' goal positions are calculated by forward kinematics. In the VR environment, the character can be replaced with the Unitree H1\cite{UnitreeH1_2021}, enabling an immersive and realistic human-robot interaction experience. Furthermore, our approach holds potential for future deployment on physical robots.
\begin{figure*}[t]
  \centering
  \begin{subfigure}{0.49\linewidth}
    \centering
    \includegraphics[width=1.0\linewidth]{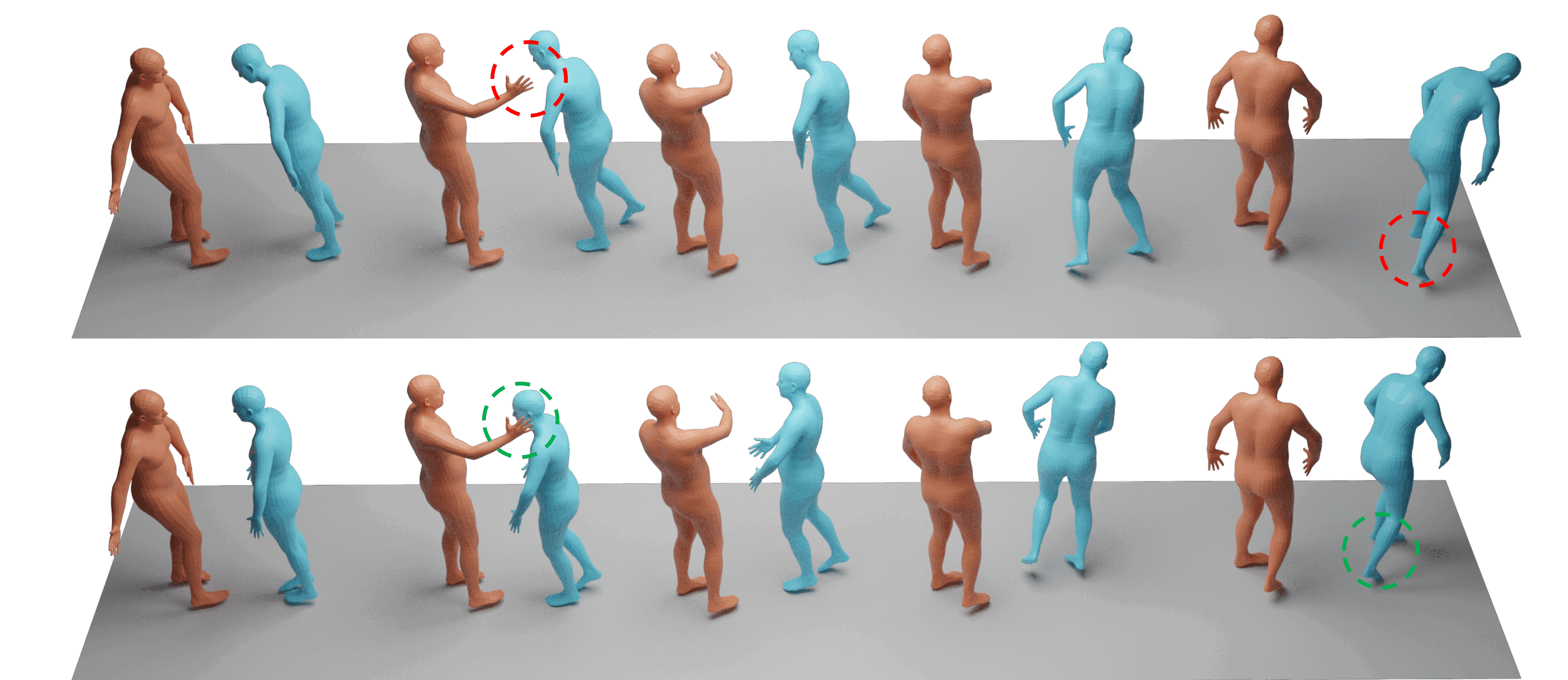}
    \label{fig:compare0}
  \end{subfigure}
  \hfill
  \begin{subfigure}{0.49\linewidth}
    \centering
    \includegraphics[width=1.0\linewidth]{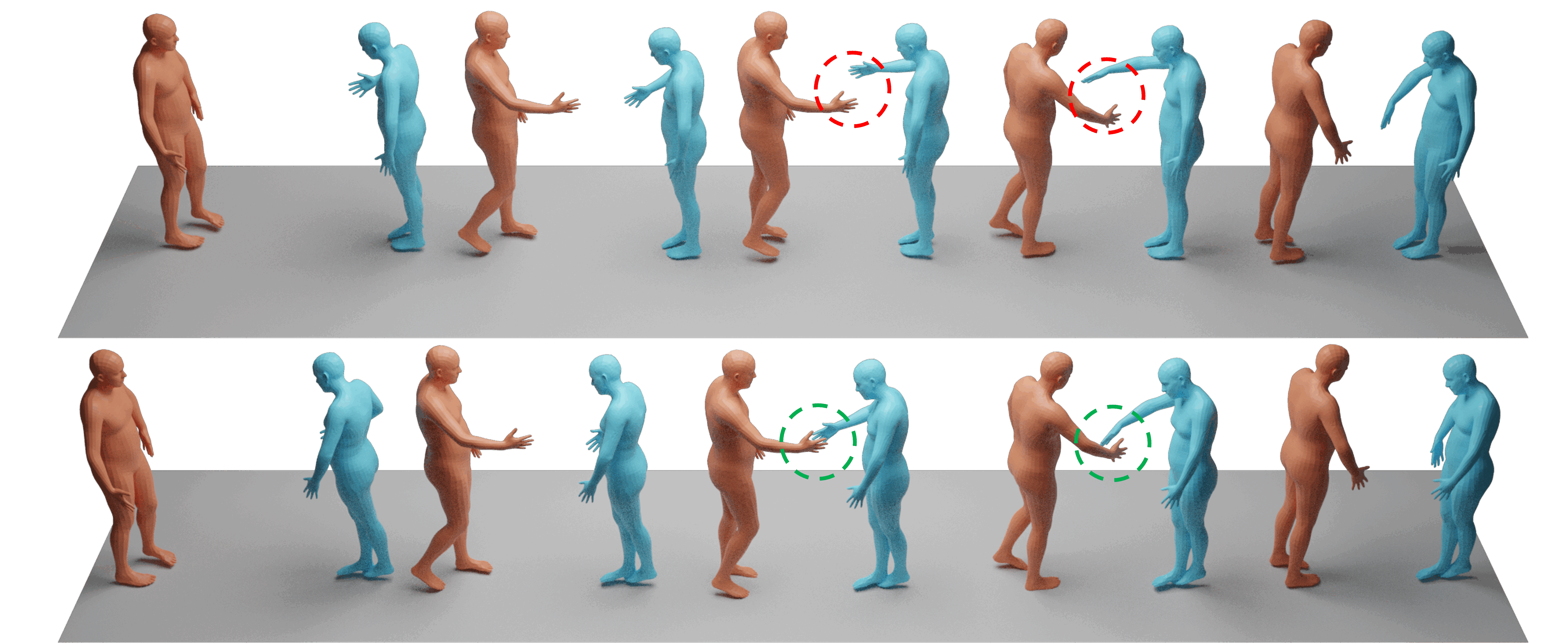}
    \label{fig:compare1}
  \end{subfigure}
  \caption{Compared to CAMDM (top row), Human-X (bottom row) achieves more complete hand contact in tasks such as face-hitting and handshaking. Additionally, its foot movement appears more natural, as highlighted in the red and green circles.}
  \label{fig:compare}
\end{figure*}

\begin{figure*}[t]
  \centering
    \includegraphics[width=1.0\linewidth]{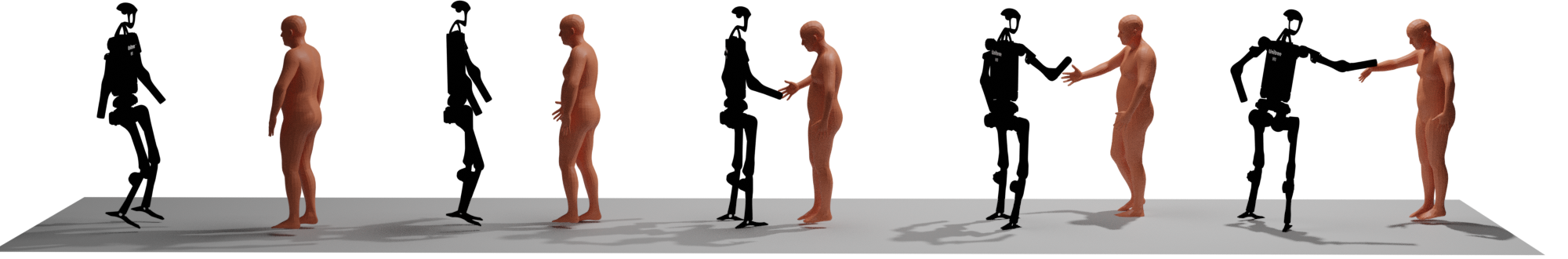}

   \caption{Visualization results on Human-Robot Interaction. The robot (black skeleton) and human (orange mesh) perform a handshake on a flat plane, from arm extension and palm contact through to shake completion, illustrating our method’s spatial coordination and motion coherence.}
   \label{fig:human-robot}
\end{figure*}

\section{Experiments}
\label{sec:exp}

\subsection{Datasets and Evaluation Metrics}

\paragraph{Datasets.} 
We evaluate our model on two datasets: Inter-X \cite{xu2024inter}, which contains 11,388 interaction SMPL-X \cite{pavlakos2019expressive} motion sequences on 40 interaction categories; InterHuman \cite{liang2024intergen}, which consists of 6022 interaction SMPL \cite{loper2015smpl} motion sequences with around 1.7M frames. For data processing, we use the actor-reactor annotations provided by \cite{xu2024regennet}. We canonicalize all the motions into the representation stated in Sec.\ref{subsubsec:rep} and set frame rate to 30 FPS.

\paragraph{Evaluation Metrics.} We assess the generated reactor's sequence using three key aspects: \textbf{1)} the quality of the reactor's motion independent of the actor, following kinematic motion generation works \cite{guo2022generating, tevet2023human}; \textbf{2)} the physical plausibility of the reactor's motion, based on physics-based methods \cite{yuan2023physdiff, tevet2024closd}; and \textbf{3)} the interaction quality between the actor and reactor, as discussed in \cite{liu2024physreaction, siyao2024duolando}.

For \textbf{1)}, we evaluate using \textit{Frechet Inception Distance (FID)} \cite{heusel2017gans} to measure the similarity between the generated and ground truth data distributions, \textit{Diversity} to assess motion variability, and \textit{Multi-modal Distance (MMDist)} to measure the distance between the generated reaction and the text prompt. For \textbf{2)}, we assess \textit{Penetration}, the average distance between the ground and the lowest body joint position below the ground, \textit{Floating}, the average distance between the ground and the lowest body joint above the ground, and \textit{Skating}, the average horizontal displacement during frames exhibiting skating behavior. For \textbf{3)}, we measure maximum \textit{Interpenetration Volume (IV)} \cite{liu2024physreaction} of reactor vertices intersecting the actor mesh, and cross-distance metrics $\textit{FID}_{cd}$ and $\textit{Div}_{cd}$ in a 100-dimensional cross-distance space \cite{siyao2024duolando} to evaluate interaction realism. Details of metrics calculation in Appendix \ref{sup:metrics}.

\subsection{Implementation Details}
In our auto-regressive denoiser, we adopt the DIT architecture \cite{peebles2023scalable} consisting of 8 Transformer layers with a 512-dimensional hidden size. We employ a frozen DistilBERT encoder~\cite{devlin2019bert, sanh2019distilbert} as the text encoder, which is widely used in motion generation tasks~\cite{guo2022generating, tevet2024closd}. For training and inference, we set the binary interaction field $\mathcal{I} \in \mathbb{R}^{6 \times 6}$, the diffusion timestep $T = 8$, the classifier-free guidance weight $w = 5$, and the foot-contact, interaction, and prefix loss weights to $\lambda_{\text{foot}} = 0.2$, $\lambda_{\text{inter}} = 0.5$, and $\lambda_{\text{prefix}} = 0.1$, respectively. In the auto-regressive setting, we use an interaction history length of $h = 20$ frames and a prediction window length of $k = 40$ frames. We train for 300K denoising steps on both the InterHuman and Inter-X datasets, and observe optimal checkpoints at 250K steps for InterHuman and 220K steps for Inter-X. All models are trained on a single \textit{NVIDIA GeForce RTX 4090d} GPU for approximately 30 hours.

For physics policy training, we use the multiplicative version of PHC network with three primitives, and construct task goal with only target body keypoints. The policy is also trained on the same device and datasets as the former stage for 20 hours. The tracking policy is executed at 30 Hz, while physics simulation is processed in NVIDIA's Isaac Gym\cite{makoviychuk2021isaac} at 60 Hz. In the starting phase of inference, we capture the first 1 second motion from the actor, and store it in the buffer along with masked reactor motion prefix for the next planning horizon's generation, the downstream reactor tracker receives default joint positions as tracking goals. After warming-up, the diffusion planner generates reasonable action-reaction motion sequences, the reactor is able to continuously follow the planning and receives both real and generated actor sequences for policy correction.

\subsection{Actor-To-Reactor Evaluation}

\paragraph{Baselines.} To evaluate the performance of our method, we adopt and reproduce several human reaction generation models as baselines: 1) InterFormer 
\cite{chopin2023interaction}: a transformer-based model with temporal and spatial attention for human reaction generation, 2) InterGen \cite{liang2024intergen}: a diffusion-based model for generating human interaction motions conditioned on text prompts., 3) ReGenNet \cite{xu2024regennet}: a diffusion-based model for generating human reactions conditioned on given human actions. and 4) CAMDM \cite{chen2024taming}: a transformer-based Conditional Auto-regressive Motion Diffusion Model. However, since it is designed for single-person motion generation, we utilize the action as a condition input to generate the corresponding reaction.

\paragraph{Comparisons.} As demonstrated in Tab. \ref{tab:interx}, in the online unconstrained reaction setting, our method exhibits significant improvements over baselines in Reaction Quality, Physical Plausibility, and Interaction Quality on both the Inter-X datasets. Unlike InterFormer's transformer architecture, our diffusion-based approach effectively captures the nuanced structure of data distributions, resulting in reactor motions that closely align with real-world motion patterns and thereby enhancing reaction quality. In comparison to InterGen and ReGenNet, our implementation of Auto-regressive Motion Diffusion Models, augmented with a Physics Tracker, enables real-time interaction with the reactor by predicting future frames. This ensures that the generated motions adhere to fundamental physical laws, producing more natural and believable animations. Furthermore, relative to CAMDM, we introduce an Interaction Loss, leading to substantial improvements in IV and $\text{FID}_{cd}$, thereby facilitating more natural interactions. The experiments on the InterHuman dataset can be found in \ref{sup:interhuman}.


\begin{table*}[htbp]
    \vspace{-1mm}
    \begin{center}
    \resizebox{1.0\textwidth}{!}{
        \begin{tabular}{l  c c c || c c c || c c c c}
        \toprule
        \multirow{2}{*}{Methods}  & \multicolumn{3}{c}{Reaction} & \multicolumn{3}{c}{Physics} & \multicolumn{3}{c}{Interaction} \\
        \cmidrule(rl){2-10} 
        & FID$\downarrow$ & Div.$\rightarrow$ & MMDist. $\downarrow$ & Pene. $\downarrow$ & Skate. $\downarrow$ & Float. $\downarrow$ & IV $\downarrow$ & $\text{FID}_{cd}$ $\downarrow$ & $\text{Div}_{cd}$ $\rightarrow$ 
        \\
        \midrule
        Ground Truth & 0.002 & 6.028 & 3.524 & 0.000 & 0.023 & 7.956 & 0.024 & 0.235 & 11.471\\
        \midrule
        InterFormer \cite{chopin2023interaction} & 6.150 & 4.922 & 8.648 & 0.524 & 0.175 & 24.109 & 0.332 & 6.825 & 8.214  \\
        InterGen \cite{liang2024intergen} & 5.485 & 4.417 & 6.389 & 0.270 & 0.143 & 11.915 & 0.269 & 3.437 & 8.730  \\
        ReGenNet \cite{xu2024regennet} & 2.212 & 5.047 & 5.235 & 0.141 & 0.118 & 9.810 & 0.218 & 1.950 & 9.248  \\
        CAMDM \cite{chen2024taming} & 1.430 & 5.657 & 5.318 & 0.153 & 0.127 & 9.864 & 0.145 & 2.086 & 9.017  \\
        \midrule
        \rowcolor{gray!20}
        Human-X & \textbf{0.975} & \textbf{6.063} & \textbf{4.115} & 
        0.118 & 
        0.092 &
        8.650 &
        0.076 & \textbf{1.692} & \textbf{9.735}  \\

        \rowcolor{gray!20}
        Human-X* & \text{1.220} &
        5.140 & 
        5.076 & 
        \textbf{0.026} & \textbf{0.007} & \textbf{4.342} & \textbf{0.054} & 1.833 & 8.565  \\
        \bottomrule
        \bottomrule
        \end{tabular}
    }
    \end{center}
    \vspace{-4mm}
    \caption[caption]{\textbf{Action-to-Reaction} with online unconstrained reaction setting on \textbf{Inter-X} \cite{xu2024inter} dataset. A higher or lower value is better for $\uparrow$ or $\downarrow$, and $\rightarrow$ means the value closer to ground truth is better. * denotes the method with reaction policy designed for humanoid reaction.}
    \label{tab:interx}
\end{table*}

\subsection{Ablation Study}
\begin{table*}[htbp]
    \vspace{-1mm}
    \begin{center}
    \resizebox{1.0\textwidth}{!}{
        \begin{tabular}{l l  c c c || c c c || c c c || c}
        \toprule
        \multirow{2}{*}{Class} & \multirow{2}{*}{Settings}  & \multicolumn{3}{c}{Reaction} & \multicolumn{3}{c}{Physics} & \multicolumn{3}{c}{Interaction} & Latency \\
        \cmidrule(rl){3-11} 
        & & FID$\downarrow$ & Div.$\rightarrow$ & MMDist. $\rightarrow$ & Penetration $\downarrow$ & Skating $\downarrow$ & Floating $\downarrow$ & IV $\downarrow$ & $\text{FID}_{cd}$ $\downarrow$ & $\text{Div}_{cd}$ $\rightarrow$ & (ms)
        \\
        \midrule
        & Ground Truth & 0.002  & 6.028  & 3.524  & 0.000  & 0.023  & 7.956  & 0.024  & 0.235  & 11.471 & -\\
        \midrule
        \multirow{3}{*}{Losses} 
        & w.o. $\mathcal{L}_{inter}$ & 1.457 & 6.154 & 4.264 & 0.134 & 0.107 & 8.715 & 0.119 & 2.208 & 9.917  & 13.6\\
        & w.o. $\mathcal{L}_{foot}$  & 1.009 & 6.114 & 4.205 & 0.127 & 0.239 & 9.120 & 0.094 & 1.687 & 9.849  & 13.6\\
        & w.o. $\mathcal{L}_{prefix}$& 1.023 & 6.128 & 4.225 & 0.133 & 0.113 & 8.857 & 0.100 & 1.697 & 9.865  & 13.6\\
        \midrule
        \multirow{4}{*}{Timesteps} 
        & 2   & 1.809 & 5.517 & 5.007 & 0.144 & 0.113 & 9.008 & 0.153 & 2.008 & 7.513 & 3.8\\
        & 5   & 1.107 & 6.263 & 4.514 & 0.128 & 0.108 & 8.908 & 0.115 & 1.807 & 8.003 & 8.2\\
        & 10  & 0.988 & 6.008 & 4.208 & 0.118 & 0.096 & 8.709 & 0.097 & 1.684 & \textbf{9.853}  & 16.9\\
        & 100 & \textbf{0.908} & 5.907 & 3.963 & 0.113 & 0.093 & 8.608 & 0.087 & 1.610 & 9.708  & 170\\
        \midrule
        \multirow{4}{*}{Horizon} 
        & $h=10, k=30$ & 1.055 & 6.106 & 4.107 & 0.123 & 0.103 & 8.805 & 0.103 & 1.687 & 9.807  & 9.6\\
        & $h=15, k=30$ & 0.987 & 6.005 & 4.006 & 0.118 & 0.097 & 8.707 & 0.098 & 1.667 & 9.787  & 10.1\\
        & $h=20, k=20$ & 0.957 & 5.956 & 3.958 & 0.113 & 0.094 & 8.658 & 0.093 & 1.626 & 9.756  & 9.5\\
        & $h=15, k=45$ & 1.006 & 6.054 & 4.105 & 0.123 & 0.104 & 8.806 & 0.102 & 1.688 & 9.809  & 13.6\\
        \midrule
        \rowcolor{gray!20}
        \multirow{2}{*}{} 
        & text & 0.926 & 5.951 & \textbf{3.909} & \textbf{0.112} & \textbf{0.087} & \textbf{8.218} & 0.092 & \textbf{1.607} & 9.822 & 13.6\\
        \rowcolor{gray!20}
        & unconstrained &0.975 & \textbf{6.063} & 4.115 & 0.118 & 0.092 & 8.650 & \textbf{0.076} & 1.692 & 9.735 & 13.6\\
        \bottomrule
        \end{tabular}
    }
    \end{center}
    \vspace{-4mm}
    \caption[caption]{\textbf{Ablation studies} of online reaction setting on the \textbf{Inter-X} \cite{xu2024inter} dataset.}
    \label{tab:ablations}
\end{table*}
We primarily conduct ablation studies on three aspects: Losses, Timesteps, and Horizon. For more experiments, please refer to Section \ref{sup:ablation}.

\paragraph{Losses.} To verify the effectiveness of our designed loss function, we ablate $\mathcal{L}_{inter}$, $\mathcal{L}_{foot}$, and $\mathcal{L}_{prefix}$ separately. As shown in Tab. \ref{tab:ablations}, removing any of them degrades performance. Notably, excluding $\mathcal{L}_{inter}$ significantly increases $\textit{FID}$ and $\textit{FID}_{cd}$, underscoring its importance for generating realistic interactions. Removing $\mathcal{L}_{foot}$ raises Skating and Floating scores, showing its role in reducing foot sliding and enhancing physical plausibility.


\paragraph{Timesteps.} During the denoising process, we experiment with different sampling timesteps: 2, 5, 10, and 100. As shown in the Tab. \ref{tab:ablations}, increasing the number of timesteps improves the model's performance; however, it also leads to higher latency. When timesteps = 10, the model has already achieved satisfactory results. Considering both generation quality and latency, we ultimately select timesteps = 8 as a trade-off, ensuring high-quality generation while meeting real-time requirements.

\paragraph{Horizon.} During inference process, the model takes a continuous sequence of h history frames as input to predict k future frames. We conduct experiments with different values of h and k, and the results in Tab. \ref{tab:ablations} show that the best performance is achieved when h = 20 and k = 40.
\section{Conclusion}
\label{sec:conclusion}
We present Human-X Interaction, a novel framework for real-time human interaction synthesis that seamlessly integrates autoregressive diffusion modeling with an actor-aware reaction policy. Our approach addresses the critical challenges of real-time synchronization, physical realism, and safety in human-avatar, human-humanoid and human-robot interactions. Specifically, Human-X jointly predicts actions and reactions using both partner’s motion histories, ensuring immersive and realistic motion synthesis. By incorporating comprehensive interaction-aware rewards and training the tracking policy with reinforcement learning, we achieve physically feasible and artifact-free interactions even in dynamic and safety-critical scenarios. Extensive experiments on the Inter-X and InterHuman datasets demonstrate significant improvements in motion quality, interaction continuity, and physical accuracy over existing methods. The current formulation focuses on synthesizing reactive motions and does not account for collaborative task completion between humans and machines. Extending the framework to support joint task execution with shared goals remains an open challenge.

\section{Acknowledgement}
\label{sec:acknowledgement}
This work was supported by Shanghai Local College Capacity Building Program(23010503100), NSFC (No.62406195), Shanghai Frontiers Science Center of Human-centered Artificial Intelligence (ShangHAI), MoE Key Laboratory of Intelligent Perception and Human-Machine Collaboration (ShanghaiTech University), HPC Platform and Core Facility Platform of Computer Science and Communication of ShanghaiTech University and Shanghai Engineering Research Center of Intelligent Vision and Imaging.


{
    \small
    \bibliographystyle{ieeenat_fullname}
    \bibliography{main}
}

\clearpage
\setcounter{page}{1}
\maketitlesupplementary

\appendix{   
    \hypersetup{linkcolor=black}
    \begin{Large}
        \textbf{Appendix}
    \end{Large}
    \etocdepthtag.toc{mtappendix}
    \etocsettagdepth{mtchapter}{none}
    \etocsettagdepth{mtappendix}{subsection}
    \newlength\tocrulewidth
    \setlength{\tocrulewidth}{1.5pt}
    \parindent=0em
    \etocsettocstyle{\vskip0.5\baselineskip}{}
    \tableofcontents
}

\section{Overview}
\label{sup:overview}
In this document, we provide additional technical details, extended experimental results, and further discussions that complement and elaborate upon the material in the main paper. Specifically, Section \ref{sup:implementation} offers a detailed description of our implementation, covering the training of both the motion diffusion model and the tracking policy, as well as clarifications on reproducing the baseline methods. In Section \ref{sup:exp}, we present additional experiments—including ablation studies—and describe the evaluation metrics in greater depth. We also explore how our approach generalizes to other settings, illustrate further results with additional visualizations, and discuss findings from our user study. Finally, Section \ref{sup:limit} lists extended failure cases of our method and offers insight into potential directions for future research. Through this supplementary material, we aim to provide a more comprehensive view of our approach, offer clarity on the nuances of the methodology, and furnish evidence of its robustness and versatility.

\section{Supplementary Implementation Details}
\label{sup:implementation}

\subsection{Details of Motion Diffusion Training}
\label{sup:train}
\paragraph{Scheduled training.}
To enhance stability and generalization of auto-regressive models, we use the scheduled training strategy \cite{bengio2015scheduled,rempe2021humor,yuan2023learning,zhao2024dart} to progressively supervise the training process with current sampling distribution. The interaction denoising network is trained on sequences of $N$ interaction windows by integrating its own full-sampling predictions into the interaction history. Instead of always relying on the ground-truth data for past motion sequences, we substitute in the model's own earlier prediction. This approach let the model encounter an inference-time distribution, such as previously unseen interactions or unconventional combinations of actor states and text prompts. 

To ease the transition to these more challenging scenarios, we adopt a three-phase training schedule:
\begin{itemize}
    \item \textbf{Fully Supervised Phase}. The model is initially trained using only ground-truth motion history.
    \item \textbf{Mixed Training Phase} The ground-truth history is gradually replaced by rollout history with a probability that increases linearly from 0 to 1, allowing the model to slowly adjust to relying on its own outputs.
    \item \textbf{Rollout Training Phase} Finally, the model is trained exclusively using the sampled inteaction history.
\end{itemize}
In practice, we train each stage for 100K iteration steps and set the consecutive window number $N=3$. The training algorithm is shown in Alg. \ref{alg:train}.

\begin{algorithm}[ht]
\caption{Scheduled training for auto-regressive interaction diffusion}
\begin{algorithmic}[1]
\State \textbf{Input:} denoiser $\mathcal{G_\theta}$ with parameters $\theta$, reactor dataloader $\mathcal{X}$, actor dataloader $\mathcal{Y}$, text dataloader $\mathcal{C}$, total diffusion steps $T$, consecutive window number $N$, optimizer $\mathcal{O}$, training loss $\mathcal{L}$, max iteration $I_{max}$, auto-regressive interaction sampler $\mathcal{S}$.
\State $iter \gets 0$

\While{$iter < I_{max}$}
    \State $[\mathbf{x}^0, \mathbf{x}^1,...,\mathbf{x}^N] \sim \mathcal{X}, [\mathbf{y}^0, \mathbf{y}^1,...,\mathbf{y}^N] \sim \mathcal{Y}$\\
    {\color{gray}\Comment{sample $N$ interaction windows from dataset} }
    \State $\mathcal{Z} \gets \{\Call{Canonicalize}{\mathbf{x}^{0}, \mathbf{y}^{0}}\}$\\
    {\color{gray}\Comment{initialize interaction history}}
    \For{$i \gets 1$ \textbf{to} $N$} {\color{gray} \Comment{number of rollouts}}
        \State $\mathbf{z}_0^i\gets \Call{Canonicalize}{\mathbf{x}^{i}, \mathbf{y}^{i}}$
        \State $t \sim \mathcal{U}[0, T)$
        \State $\mathbf{z}_t^i \gets \Call{Forward\_diffusion}{\mathbf{z}_0^i, t}$
        \State $c^i \gets \Call{Collect}{\mathcal{C}}, \mathbf{z}^{i-1} \gets \Call{Collect}{\mathcal{Z}}$
        \State $\hat{\mathbf{z}}_0^i = \mathcal{G}_{\theta}(\mathbf{z}_t^i, \mathbf{z}^{i-1},t, c^i$)  \\
        {\color{gray} \Comment{next interaction denoising}}
        \State $\nabla \gets \nabla_{\theta}  \mathcal{L}(\mathbf{z}_0^i, \hat{\mathbf{z}}_0^i, \mathbf{z}^{i-1})$  
        \State $ \theta \gets \mathcal{O}(\theta, \nabla)$ {\color{gray} \Comment{back propagation}}
        \State $p \gets$ \Call{Schedule\_probability}{$iter$, $I_{max}$}  \\
        {\color{gray} \Comment{sample schedule training probability}}
        \If{$\text{rand()} > p$}
            \State $\mathbf{z}_T^i \gets \Call{Forward\_diffusion}{\mathbf{z}_0^i, T}$ \\
            {\color{gray} \Comment{maximum noising} }
            \State $\hat{\mathbf{z}}_0^i = \mathcal{S}(\mathcal{G}_{\theta}, \mathbf{z}_T^i,  \mathbf{z}^{i-1}, T, c^i)$  \\
            {\color{gray}\Comment{full sampling loop}}
            \State $\hat{\mathbf{x}}_0^i = \Call{Recover}{\hat{\mathbf{z}}_0^i}$ 
            \State $\tilde{\mathbf{z}}^i_0 = \Call{Canonicalize}{\hat{\mathbf{x}}_0^i, \mathbf{y}^{i}}$
            \State $\mathcal{Z} \gets \mathcal{Z} \cup \tilde{\mathbf{z}}_0^i$ \\
            {\color{gray}\Comment{set predicted reaction into history}}
        \Else
            \State $\mathcal{Z} \gets \mathcal{Z} \cup \mathbf{z}_0^i$ \\
            {\color{gray}\Comment{use dataset interaction history}}
        \EndIf
        \State $iter \gets iter + 1$
    \EndFor
\EndWhile
\end{algorithmic}
\label{alg:train}
\end{algorithm}

\subsection{Details of Reaction Policy Training}
\label{sup:policy-train}
\paragraph{Actor-aware Reaction Policy} Policy observations include generated and captured actor motion to adjust original imitation rewards and prevent interpenetration.
The distances of every joint's positions are summed to get the adaptive weight $w( \hat{y}, y_{real})$, 
we use a linear interpolate between goal rewards depending on synthesis and realistic data. The rewards based on real action $y_{real}$ are the imitation reward at default joint position and root distance reward up to 1, these rewards tend to keep the reactor away from the unpredictable actor,
\begin{equation}
    \begin{aligned}
        r(s,a,& s', \hat{x}, \hat{y}, y_{real}) = (1-w( \hat{y}, y_{real})) * r^{\text{PHC}}(s,a, \hat{x})\\
        & + w( \hat{y}, y_{real}) * (r^{\text{default}}(s, a) + r^{\text{root}}(s, y_{real})),
    \end{aligned}
\end{equation}
where $r^{\text{PHC}}(s^n, \hat{x}^n)$ is the reward components of PHC in full-motion imitation. 

The imitation reward $r^{\text{PHC}}$ proposed by PHC refers to $0.5 r^{g} + 0.5 r^{\text{amp}} + r^{\text{energy}}$. The weight $w(\hat{y}, y_{real})$ is derived from the cosine similarity $S_C$ of 24 SMPL joints at frame $i$, then:
\begin{equation}
    w(\hat{y}, y_{real}) = \frac{1}{2}(
        1-\frac{1}{T}\sum\nolimits_{i=1}^{T}S_C(\hat{y_i}, y^{real}_i)
    )
\end{equation}
Once the prediction and capture conflict, $w$ increases and the policy prioritizes higher rewards from the terms $r^{\text{defualt}} + r^{\text{root}}$. $r^{\text{defualt}}=0.5e^{-100 \|p^{\overline{x}}_i - p_i\|}$, representing the imitation reward with the average standing or walking motion $\overline{x}$ as its goal
($p$: joint position).
While $r^{\text{root}}$ penalizes root proximity to the actor, with no reward given beyond 0.4 meters: $\text{max}(\|p^{real}_{root,i} - p_{root,i}\|, 0.4)$. All the above process based on the precise capture for real actors. For noisy captures, we add slight noise to the real actor motion during policy training and simulate huge observation noises via mismatching generated actor and captured actor sequences. While domain randomization cannot fully erase every possible noise without additional observation modalities (e.g., inertial measurement units).

\section{Additional Experiments and Results}
\label{sup:exp}

\subsection{Evaluations on InterHuman dataset}
\label{sup:interhuman}
\begin{table*}[htbp]
    \vspace{-1mm}
    \begin{center}
    \resizebox{1.0\textwidth}{!}{
        \begin{tabular}{l  c c c || c c c || c c c }
        \toprule
        \multirow{2}{*}{Methods}  & \multicolumn{3}{c}{Reaction} & \multicolumn{3}{c}{Physics} & \multicolumn{3}{c}{Interaction} \\
        \cmidrule(rl){2-10} 
        & FID$\downarrow$ & Div.$\rightarrow$ & MMDist. $\downarrow$ & Pene. $\downarrow$ & Skat.$\downarrow$ & Float. $\downarrow$ & IV $\downarrow$ & $\text{FID}_{cd}$ $\downarrow$ & $\text{Div}_{cd}$ $\rightarrow$ 
        \\
        \midrule
        Ground Truth  & 0.008 & 3.865 & 4.306 & 0.000 & 0.032 & 21.273 & 0.045 & 0.532 & 9.109 \\
        \midrule
        InterFormer \cite{chopin2023interaction} & 9.475 & 2.645& 10.893 & 1.113 & 1.021 & 37.927 & 0.354 & 5.734 & 5.302  \\
        InterGen \cite{liang2024intergen} &  6.379 & 3.033 & 7.881 & 0.266 & 0.279 & 23.398 & 0.210 & 3.101 & 6.414  \\
        ReGenNet \cite{xu2024regennet} & 2.257 & 3.459 & 5.754 & 0.427 & 0.396 & 24.804 & 0.169 & 1.733 & 7.485  \\
        CAMDM \cite{chen2024taming} & 2.166 & 4.161 & 6.212 & 0.296 & 0.177 & 24.261 & 0.341 & 3.701 & \textbf{8.269}  \\
        \midrule
        \rowcolor{gray!20}
        Human-X & \textbf{1.995} & \textbf{3.767} & \textbf{5.638} & 0.216 & 0.048 & 22.165 & \textbf{0.106} & \textbf{1.582} & 7.754  \\
        \rowcolor{gray!20}
        Human-X* & 2.350 & 3.245 & 6.071 & \textbf{0.064} & \textbf{0.008} & \textbf{12.026} & 0.134 & 1.634 & 7.015  \\
        \bottomrule
        \end{tabular}
    }
    \end{center}
    \vspace{-4mm}
    \caption[caption]{\textbf{Action-to-Reaction} with online unconstrained reaction setting on \textbf{InterHuman}\cite{liang2024intergen} dataset. A higher or lower value is better for $\uparrow$ or $\downarrow$, and $\rightarrow$ means the value closer to ground truth is better. * denotes the method with the physics tracker}
    \label{tab:interhuman}
\end{table*}
On the InterHuman dataset as shown in Tab \ref{tab:interhuman}, our method shows a slight decline in $\text{Div}_{cd}$ compared to CAMDM. We attribute this to the actor's movements constraining the reactor's actions: the incorporation of Interaction Loss reduces the distance between the actor and reactor to encourage contact. In contrast, CAMDM treats the actor as a conditioning factor, imposing fewer restrictions on the reactor's movements. More details about experiments within online text-guided reaction setting can be found in \ref{sup:ablation}.

\subsection{Details of Evaluation Metrics}
\label{sup:metrics}
The evaluation framework for synthesized reactor sequences encompasses three principal dimensions: (1) FID, Diversity and Multimodal Distance (MMDist) for Reaction Quality, which assesses the reactor’s motion independently of the actor, following\cite{guo2022generating, tevet2023human}; (2) Penetration, Floating and Skating for Physical Plausibility, which evaluates the adherence of the reactor’s motion to physical constraints, following\cite{yuan2023physdiff, tevet2024closd}; and (3) Interpenetration Volume (IV), $\text{FID}_{cd}$ and $\text{Div}_{cd}$ for Interaction Quality which examines the realism of interactions between the actor and the reactor, following\cite{liu2024physreaction, siyao2024duolando}. All the motion and text features are extracted with the pretrained checkpoints in \cite{guo2022generating}. The metrics are defined in detail as follows.

\paragraph{FID.}Frechet Inception Distance (FID) is a widely adopted metric for quantifying the quality of generated motion by measuring the statistical divergence between feature distributions of real and synthesized samples. It evaluates the fidelity of synthesized motion sequences by comparing their latent-space representations to those of ground-truth motion data.

\paragraph{Diversity.}Diversity measures the variance of action categories across all generated motion sequences. Specifically, we sample two subsets of motion sequences, each containing the same number of samples $S_d$ from the set of generated motion sequences across all action categories, denoted as $\left\{ \mathbf{v}_1, ..., \mathbf{v}_{S_d} \right\}$ and $\left\{ \mathbf{v}_1', ..., \mathbf{v}_{S_d}' \right\}$. The diversity of the generated motions is then defined as $\text{Diversity}=\frac{1}{S_d} \sum_{i=1}^{S_d} \left\| \mathbf{v}_i - \mathbf{v}'_i \right\|_2$. In our experiments, we set $S_d = 200$. As the diversity of the generated motions approaches that of the real dataset, the generated motions exhibit greater diversity, leading to improved alignment with real-world motion distributions.

\paragraph{MMDist.}Multimodal Distance (MMDist) evaluates the alignment fidelity between generated motion features and their corresponding text features by measuring the average Euclidean distance. Formally, given N paired motion-text samples, we extract motion and text features using pretrained extractors, denoted as $f_i^{motion}$ and $f_i^{text}$ respectively. The MMDist is computed as: $\text{MMDist}=\frac{1}{N} \sum_{i=1}^{N} \left\| f_i^{motion} - f_i^{text} \right\|_2$, where lower MMDist indicates that the generated motion aligns more closely with the textual description.

\paragraph{Penetration, Floating and Skating.}\textbf{Penetration} measures the distance between the lowest body mesh vertex below the ground and the ground surface, assessing whether the character exhibits ground penetration. \textbf{Floating} computes the distance between the lowest body mesh vertex above the ground and the ground surface, evaluating whether the character is unnaturally floating. \textbf{Skating} identifies foot joints that maintain ground contact across consecutive frames and calculates their average horizontal displacement, assessing the presence of foot sliding artifacts.

\paragraph{Interpenetration Volume.}Interpenetration Volume (IV) measures the collision volume between the meshes of the actor and reactor, serving as a penalty term to discourage mesh interpenetration and unintended collisions between the two entities.

\paragraph{$\text{FID}_{cd}$ and $\text{Div}_{cd}$.}In every frame, we select 10 key joints for both the actor and the reactor, chosen from the complete set of available joints. The selected joints encompass the pelvis, knees, feet, shoulders, head, and two wrists. A matrix $M \in \mathbb{R}^{10 \times 10}$ is then constructed based on the pairwise distances between the selected joints of the two agents, serving as the interactive feature. Based on this representation, we compute the $\text{FID}_{cd}$ and $\text{Div}_{cd}$, which are used to supervise and assess the interaction quality between the two characters.

\subsection{Additional Experiments on Diffusion Planner}
\label{sup:ablation}
\begin{table*}[htbp]
    \label{tab:sup-interx}
    \vspace{-1mm}
    \begin{center}
    \resizebox{1.0\textwidth}{!}{
        \begin{tabular}{l  c c c || c c c || c c c }
        \toprule
        \multirow{2}{*}{Methods}  & \multicolumn{3}{c}{Reaction} & \multicolumn{3}{c}{Physics} & \multicolumn{3}{c}{Interaction} \\
        \cmidrule(rl){2-10} 
        & FID$\downarrow$ & Div.$\rightarrow$ & MMDist. $\downarrow$ & Pene. $\downarrow$ & Skate. $\downarrow$ & Float. $\downarrow$ & IV $\downarrow$ & $\text{FID}_{cd}$ $\downarrow$ & $\text{Div}_{cd}$ $\rightarrow$ \\
        \midrule
        Ground Truth  & 0.002  & 6.028  & 3.524  & 0.000  & 0.023  & 7.956  & 0.024  & 0.235  & 11.471 \\
        \midrule
        InterGen \cite{liang2024intergen}       & 5.211  & 4.498  & 6.070  & 0.257  & 0.136  & 11.319 & 0.256  & 3.265  & 8.867 \\
        ReGenNet \cite{xu2024regennet}            & 2.101  & 5.096  & 4.973  & 0.134  & 0.112  & 9.320  & 0.207  & 1.853  & 9.359 \\
        CAMDM \cite{chen2024taming}              & 1.359  & 5.676  & 5.052  & 0.145  & 0.121  & 9.371  & 0.138  & 1.982  & 9.140 \\
        \midrule
        \rowcolor{gray!20}
        Human-X  & \textbf{0.926} & \textbf{5.951} & \textbf{3.909} & \textbf{0.112} & \textbf{0.087} & \textbf{8.218} & \textbf{0.072} & \textbf{1.607} & \textbf{9.822} \\
        \bottomrule
        \end{tabular}
    }
    \end{center}
    \vspace{-4mm}
    \caption[caption]{\textbf{Action-to-Reaction} with online text-guided reaction setting on \textbf{Inter-X} \cite{xu2024inter} dataset, where a higher or lower value is better for $\uparrow$ or $\downarrow$, and $\rightarrow$ means the value closer to ground truth is better.}
    \label{tab:sup-interx}
\end{table*}

\begin{table*}[htbp]
    \vspace{-1mm}
    \begin{center}
    \resizebox{1.0\textwidth}{!}{
        \begin{tabular}{l  c c c || c c c || c c c }
        \toprule
        \multirow{2}{*}{Methods}  & \multicolumn{3}{c}{Reaction} & \multicolumn{3}{c}{Physics} & \multicolumn{3}{c}{Interaction} \\
        \cmidrule(rl){2-10} 
        & FID$\downarrow$ & Div.$\rightarrow$ & MMDist. $\downarrow$ & Pene. $\downarrow$ & Skat. $\downarrow$ & Float.$\downarrow$ & IV $\downarrow$ & $\text{FID}_{cd}$ $\downarrow$ & $\text{Div}_{cd}$ $\rightarrow$ \\
        \midrule
        Ground Truth  & 0.008  & 3.865  & 4.306  & 0.000  & 0.032  & 21.270 & 0.045  & 0.532  & 9.109 \\
        \midrule
        InterGen \cite{liang2024intergen}       & 6.060  & 3.366  & 7.486  & 0.253  & 0.265  & 22.221 & 0.200  & 2.946  & 6.549 \\
        ReGenNet \cite{xu2024regennet}            & 2.144  & 3.621  & 5.466  & 0.406  & 0.376  & 23.560 & 0.161  & 1.646  & 7.566 \\
        CAMDM \cite{chen2024taming}              & 2.058  & 4.043  & 5.901  & 0.281  & 0.168  & 23.047 & 0.324  & 3.516  & 8.311 \\
        \midrule
        \rowcolor{gray!20}
        Human-X & \textbf{1.889} & \textbf{3.807} & \textbf{5.356} & \textbf{0.205} & \textbf{0.046} & \textbf{21.052} & \textbf{0.101} & \textbf{1.513} & \textbf{8.384} \\
        \bottomrule
        \end{tabular}
    }
    \end{center}
    \vspace{-4mm}
    \caption[caption]{\textbf{Action-to-Reaction} of online text-guided reaction setting on \textbf{InterHuman} \cite{liang2024intergen} dataset, where a higher or lower value is better for $\uparrow$ or $\downarrow$, and $\rightarrow$ means the value closer to ground truth is better.}
    \label{tab:sup-interhuman}
\end{table*}
\begin{table*}[htbp]
    \vspace{-1mm}
    \begin{center}
    \resizebox{1.0\textwidth}{!}{
        \begin{tabular}{l l  c c c || c c c || c c c  || c}
        \toprule
        \multirow{2}{*}{Class} & \multirow{2}{*}{Settings}  & \multicolumn{3}{c}{Reaction} & \multicolumn{3}{c}{Physics} & \multicolumn{3}{c}{Interaction} & Latency \\
        \cmidrule(rl){3-11} 
        & & FID$\downarrow$ & Div.$\rightarrow$ & MMDist. $\downarrow$ & Pene. $\downarrow$ & Skat. $\downarrow$ & Float. $\downarrow$ & IV $\downarrow$ & $\text{FID}_{cd}$ $\downarrow$ & $\text{Div}_{cd}$ $\rightarrow$ & (ms)
        \\
        \midrule
         & Ground Truth & 0.002  & 6.028  & 3.524  & 0.000  & 0.023  & 7.956  & 0.024  & 0.235  & 11.471 & - \\
        \midrule
        \multirow{4}{*}{Representation} 
        & non-canonical                    & 2.032 & 6.285 & 4.278 & 0.129 & 0.131 & 8.814 & 0.083 & 5.737 & 7.853  & 10.2\\[0.5ex]
        & w.o. $\mathcal{I}$              & 1.108 & 5.842 & 4.235 & 0.121 & 0.095 & 8.689 & 0.078 & 3.542 & 8.423  & 13.1\\[0.5ex]
        & w.o. $\dot{\mathbf{r}}^y$ and $\dot{\mathbf{p}}^y$ & 1.044 & 5.617 & 4.259 & 0.125 & 0.094 & 8.712 & 0.080 & 2.389 & 9.168  & 12.8\\[0.5ex]
        & add $ \theta^y$                 & 1.363 & 5.956 & 4.116 & 0.117 & 0.089 & 8.634 & 0.075 & 1.864 & 9.705  & 15.0\\
        \midrule
        \multirow{3}{*}{$\mathcal{I}$ Field Size} 
        & $1 \times 1$                   & 1.095 & 5.945 & 4.305 & 0.133 & 0.102 & 8.916 & 0.087 & 1.802 & 9.223  & 11.9\\[0.5ex]
        & $10 \times 10$                 & 0.979 & 6.092 & 4.153 & 0.119 & 0.094 & 8.667 & 0.077 & 1.701 & \textbf{9.781}  & 13.9\\[0.5ex]
        & $22 \times 22$                 & 0.982 & 6.067 & 4.118 & 0.121 & \textbf{0.089} & 8.650 & \textbf{0.069} & 1.697 & 9.399  & 15.1\\
        \midrule
        \multirow{3}{*}{Num of $l_{layers}$} 
        & 2                              & 2.729 & 5.611 & 4.222 & 0.124 & 0.096 & 8.701 & 0.079 & 1.827 & 7.952  & 3.8\\[0.5ex]
        & 4                              & 1.379 & 5.868 & 4.117 & 0.118 & 0.093 & 8.654 & 0.076 & 1.734 & 9.538  & 7.2\\[0.5ex]
        & 16                             & 1.007 & 5.935 & 4.183 & 0.121 & 0.094 & 8.677 & 0.078 & 1.705 & 9.654  & 25.5\\
        \midrule
        Text Encoder & CLIP\cite{radford2021learning} & 1.023 & 6.100 & \textbf{4.073} & 0.120 & 0.095 & 8.680 & 0.079 & 1.712 & 9.652  & 13.6\\
        \midrule     
        \rowcolor{gray!20}
        & Human-X                         & \textbf{0.975} & \textbf{6.063} & 4.115 & \textbf{0.118} & 0.092 & \textbf{8.650} & 0.076 & \textbf{1.694} & 9.735  & 13.6\\
        \bottomrule
        \end{tabular}
    }
    \end{center}
    \vspace{-4mm}
    \caption[caption]{Additional \textbf{ablation studies} with online reaction settings on the \textbf{Inter-X}\cite{xu2024inter} dataset, where a higher or lower value is better for $\uparrow$ or $\downarrow$, and $\rightarrow$ means the value closer to ground truth is better.}
    \label{tab:sup-ablations}
\end{table*}

In the online text-guided reaction setting, we also conduct experiments on the Inter-X and InterHuman datasets, evaluating performance across three key dimensions: Reaction Quality, Physical Plausibility, and Interaction Quality, as shown in Tab. \ref{tab:sup-interx} and \ref{tab:sup-interhuman}. Unlike the experiments in the unconstrained reaction setting, we exclude InterFormer from the baselines since it does not support text-conditioned inputs. The results demonstrate that our approach outperforms previous baselines across all metrics, achieving state-of-the-art performance. Incorporating text supervision enables the generation of diverse and fine-grained human motion sequences, allowing for detailed customization based on the provided descriptions, while also leading to slight improvements in various evaluation metrics. However, it is noteworthy that even without textual input, our model already achieves strong performance. Although the inclusion of text information enhances the results, the overall improvement is relatively modest. This indicates that the model remains robust and effective even in the absence of additional textual guidance.

In additional ablation studies, we conduct experiments on four aspects: motion representation, the size of binary interaction field $\mathcal{I}^n$, numbers of the decoder layers $l_{layers}$ and text encoder. All experimental results are presented in Tab. \ref{tab:sup-ablations}.
\paragraph{Representation.} Similar to InterGen\cite{liang2024intergen}, we attempt to use a non-canonical representation for multi-person interaction motion. However, experimental results show a performance degradation across all metrics, with a particularly significant drop in interaction-related metrics. This performance degradation is primarily due to the fact that the non-canonical representation encodes features in the global coordinate system. In contrast, our approach defines the coordinate origin at the root joint of the reactor and represents motion in a relative coordinate system. This relative formulation facilitates learning the spatial relationships between the two characters, whereas the global representation makes it substantially more challenging for the model to capture these interactions effectively.

When removing binary interaction field $\mathcal{I}$ and angular velocity of the roots $\dot{\mathbf{r}}^y$, temporal difference of the local joint positions $\dot{\mathbf{p}}^y$ separately, the model's ability to learn the authenticity of motion interaction and the temporal coherence between consecutive frames decreases, which naturally leads to a degradation in generation quality. However, introducing 6D representation of the joint rotations $\theta^y$ also results in performance degradation. This is because $\theta^y$ does not contribute to the model's understanding of the interaction between the two characters; instead, the inclusion of such redundant information hinders the model's comprehension ability.

\paragraph{$\mathcal{I}$ Field Size.} The size of the binary interaction field $\mathcal{I}$ refers to the number of joints selected in the contact map. When only the pelvis joint is selected, $\mathcal{I}$ fails to be effective, as the pelvis itself rarely engages in contact. On the other hand, selecting all 22 joints achieves the best performance on most metrics but introduces significant computational latency. Therefore, we ultimately select the six most critical joints, including the pelvis, head, both ankles, and both wrists, to ensure interaction quality while minimizing computational overhead.

\paragraph{Num of $l_{layers}$.} We conduct experiments with 2, 4, 8, and 16 decoder layers and ultimately select 8 layers as the final choice, as it demonstrated the best overall performance.

\paragraph{Text Encoder.} In ablation study, we attempt to use CLIP as the text encoder. As stated in CLOSD\cite{tevet2024closd}, CLIP\cite{radford2021learning} tends to focus on understanding image descriptions. Although it outperforms DistilBERT\cite{sanh2019distilbert} in MMDist, it performs slightly worse on other metrics. Therefore, we ultimately adopt DistilBERT as the text encoder.


\subsection{Additional Experiments on Reaction Policy}
\label{sup:policy}

\begin{table}[htbp]
\centering
\resizebox{1.0\linewidth}{!}{
\begin{tabular}{l r | r r r  r}
\toprule
\multirow{1}{*}{Methods} & $\text{IV}\!\downarrow$ &$\text{Succ}\!\uparrow$ & $\mathit{E}_{\text{mpjpe}}\!\downarrow$ & $\mathit{E}_{\text{acc}}\!\downarrow$ & $\mathit{E}_{\text{vel}}\!\downarrow$ 
\\
\midrule
PHC[33]  & 4.7& 84.1\% & 47.6 & 11.7 & 9.1\\
Ours & 1.4 & \textbf{95.6}\% & \textbf{37.3} & \textbf{10.5} & \textbf{4.2}\\
Ours-safety & \textbf{0.05} & 84.0\% & 51.7 & 11.4 & 10.1\\
\bottomrule
\end{tabular}
}

\caption{Test performance of reaction policy on Inter-X dataset.
(Ours-safety indicates actor-aware policy). 
}
\label{tab:phc-cmp}
\end{table}

In this section, we present additional experimental results on the Inter-X dataset, comparing our actor-aware reaction policy with the baseline method PHC\cite{luo2023perpetual}. The results are shown in Tab. \ref{tab:phc-cmp}. 
IV stands for interpenetration volume and other metrics following PHC. During inference, we disable the inter-actor collision avoidance in simulation, otherwise, the interpenetration volume (IV) will constantly be zero. 
Our method achieves a significant reduction in interpenetration volume and a notable improvement in success rate, while maintaining comparable performance in other metrics. 
This demonstrates that our actor-aware reaction policy effectively enhances the reactor's motion quality and interaction realism.

\subsection{User Study}


\begin{table}[h]
    \centering
    \resizebox{1.0 \columnwidth}{!}{
        \begin{tabular}{cl} 
            \toprule
            Metrics & \multicolumn{1}{c}{Question} \\
            \midrule
            Diversity  &  Which interaction leads to a greater variety of reactions? \\
            Consistency & Which interaction produces more realistic reactions? \\
            Authenticity  &  Which interaction exhibits more realistic contact? \\
            \bottomrule
        \end{tabular}
    }
    \caption{Question settings for user questionnaire.}
    \label{tab:user-study}
\end{table}
In order to assess the effectiveness of the proposed method, we conduct a user study involving 15 participants. They are asked to complete a questionnaire, with the specific questions provided in Tab \ref{tab:user-study}. In the first phase of the experiment, participants are instructed to compare the video results generated by our method against those produced by \textbf{InterGen}\cite{liang2024intergen}, \textbf{ReGenNet}\cite{xu2024regennet} and the ground truth, based on three key criteria: Diversity, Consistency, and Authenticity. In the second phase, participants will engage in an immersive interaction with the avatar using a VR headset. Following this experience, they evaluate our method in comparison to \textbf{CAMDM}\cite{chen2024taming} and select the motion sequence they perceive as superior. Each participant is presented with 20 video samples and allocated an average of 30 seconds for the VR experience, ensuring sufficient exposure to the generated results before providing their final assessment.

\begin{figure}[t]
  \centering
    \includegraphics[width=1.0\columnwidth]{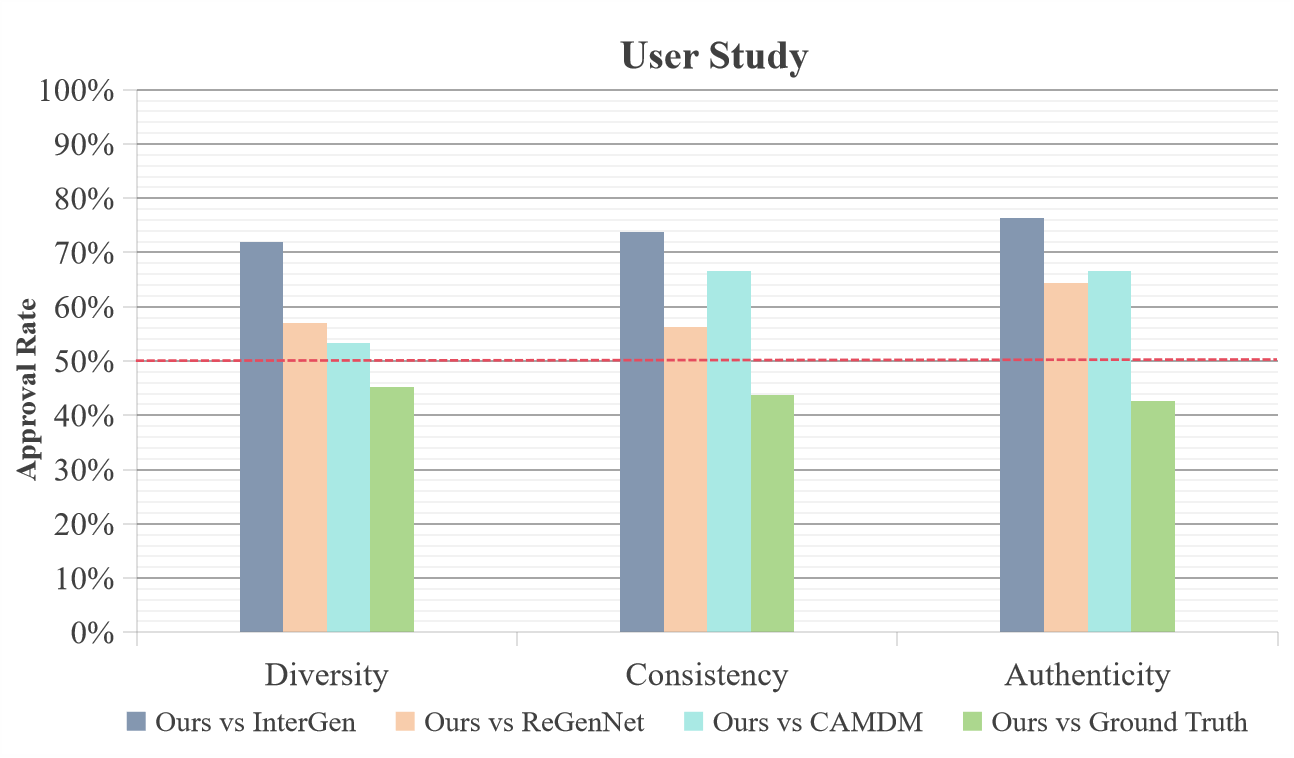}

   \caption{Extensive experimental results indicate that participants perceive our method to perform better in all three metrics: Diversity, Consistency, and Authenticity.}
   \label{fig:user-study}
\end{figure}

The final results, as presented in Figure \ref{fig:user-study}, demonstrate that our method achieved the highest approval rate across all metrics among the participants, reaching state-of-the-art performance. This result validates that our approach not only ensures latency-free performance but also maintains high-quality motion generation.

\section{Extended Limitation and Discussions}
\label{sup:limit}

\subsection{Discussion and Future Work}
In this work, we introduce a novel auto-regressive framework that seamlessly integrates predictive interaction synthesis with actor-aware physical refinement. Meanwhile, we integrate our framework into a real-time VR system, demonstrating its effectiveness in immersive, unconstrained environments. Through comparative experiments, we identify considerable areas where our model can be further improved, as outlined below:

\paragraph{Further Memory and Planning.}
Currently, our model relies on the past 20 frames to predict the next 40 frames. However, this short temporal window may result in the loss of critical historical information. A potential direction could be leveraging the planning and decision-making capabilities of Large Language Models (LLMs) to guide the model\cite{wu2024human, cai2024digital, lin2024chathuman}, allowing it to incorporate a longer temporal context for more informed decision-making and improved generation quality.

\paragraph{Multi-Modal Action and Response.}
For now, our model is limited to generating the reactor's motion based solely on the actor's motion and textual input. However, in real-world scenarios, additional modalities such as audio\cite{tseng2023edge, jiang2024solami} and visual cues\cite{driess2023palm, ho2022video} play a crucial role in motion decision-making. Developing an interactive system capable of processing multimodal inputs and outputs would enhance its generalizability and expand its potential applications.

\paragraph{Reactive Character Generalizability.}
The reactor can be made more diverse. In the future, we can extend its representation from the SMPL\cite{loper2015smpl} to SMPL-X\cite{pavlakos2019expressive} and SMPL+H\cite{MANO:SIGGRAPHASIA:2017}, enabling finer-grained control over facial expressions, hand gestures, and body shape. Additionally, the reactor can be replaced with humanoid robots\cite{chen2025rhino, chen2025symbiosim, ji2024exbody2}, laying the groundwork for real-world deployment.

\paragraph{Various Interaction Context.}
Although we have achieved real-time interaction between two individuals, real-time interaction among multiple participants presents a significantly greater challenge. While systems like \cite{cai2024digital} can generate interactions between two individuals in specific scenarios through text-based action control, they lack the capability for real-time reaction generation. Implementing real-time, two-person interactions within specific contexts, or integrating dual participants with object interaction, remains an area for future exploration.

\paragraph{Customized Reaction Design.}
In real-life scenarios, individuals possess distinct personalities, leading to varied responses to the same action. Inspired by \cite{cai2024digital}, we can assign personality traits to the reactor in the future, enabling it to generate personalized responses that cater to users' individual needs. 

\end{document}